\documentclass{article}

\PassOptionsToPackage{numbers,sort}{natbib}
\usepackage[preprint]{nips_2018}

\usepackage[T1]{fontenc}    
\usepackage[english]{babel}
\usepackage[utf8]{inputenc} 
\usepackage{algorithmicx}
\usepackage{algorithm}
\usepackage{algpseudocode}
\usepackage{amsfonts}
\usepackage{amsmath}
\usepackage{mathtools}
\usepackage{amssymb}
\usepackage{amsthm}
\usepackage{booktabs}
\usepackage{caption}
\usepackage{graphicx}
\usepackage{multicol}
\usepackage{subcaption}
\usepackage{tikz}
\usepackage{float}
\usepackage{rotating}
\usepackage{hyperref}

\DeclareMathAlphabet{\pazocal}{OMS}{zplm}{m}{n}

\theoremstyle{definition}

\title{Gradient Energy Matching \\ for Distributed Asynchronous Gradient Descent}

\author{
  Joeri R.~Hermans\\
  University of Li\`ege \\
  \texttt{joeri.hermans@doct.uliege.be}\\
  \And
  Gilles Louppe\\
  University of Li\`ege \\
  \texttt{g.louppe@uliege.be}\\
}

\begin{document}

\maketitle

\begin{abstract}
Distributed asynchronous \textsc{sgd} has become
widely used for deep learning in large-scale systems, but remains notorious for its instability when increasing the number of
workers. In this work, we study the dynamics of distributed asynchronous
\textsc{sgd} under the lens of Lagrangian mechanics. 
Using this description, we
introduce the concept of energy to describe the optimization process and derive a sufficient
condition ensuring its stability as long as the collective energy induced by the active workers
remains below the energy of a target synchronous process.
Making use of this criterion, we
derive a stable distributed asynchronous optimization procedure,
\textsc{gem}, that estimates and maintains the energy of the asynchronous system below or equal to
the energy of sequential \textsc{sgd} with momentum.
Experimental results highlight the stability and speedup of \textsc{gem} compared to existing schemes, even when scaling to one hundred asynchronous workers.
Results also indicate better generalization compared to the targeted \textsc{sgd} with momentum.


\end{abstract}


\section{Introduction}
\label{sec:introduction}

In deep learning, stochastic gradient descent (\textsc{sgd}) and its variants have become the
optimization method of choice for most training problems. For large-scale
systems, a popular variant is distributed asynchronous
optimization~\citep{dean2012large,chilimbi2014project} based on a master-slave architecture.
That is, a set of $n$ workers $w$ 
individually contribute updates $\Delta\theta_t$ asynchronously to
a master node holding the central variable $\theta_t$, under a global clock $t$, such that
\begin{equation}
  \label{eq:traditional_async_update}
  {\theta}_{t+1} = {\theta}_t + \Delta\theta_t.
\end{equation}
Due to the presence of asynchronous updates, i.e. without locks or synchronization barriers, an implicit queuing model emerges in the system~\citep{2016arXiv160509774M}, such that workers are updating ${\theta}_t$ with updates
$\Delta\theta_t$ that are possibly based on a previous parameterization
\begin{equation}
    \label{eq:staleness}
    s_t \triangleq {\theta}_{t-\tau_t}
\end{equation}
of the central variable,
where $\tau_t$ is the number
updates that occurred between the time the worker responsible for the update at time $t$ pulled (read) the central variable, and committed (wrote) its update.
The term $\tau_t$ is traditionally called the staleness or the delay of the update at time $t$. Assuming a homogeneous infrastructure, the expected staleness~\citep{2016arXiv160509774M} for a worker $w$ under a simple queuing model $\mathcal{Q}$ can be shown to be $\mathbb{E}_{\mathcal{Q}}\big[\tau_t\big] = n - 1$.
In this setup, instability issues are common because updates that are committed are
most likely based on past and outdated versions of the central variable, in particular as the number $n$ of workers increases. To mitigate
this effect, previous approaches~\citep{2015arXiv151105950Z,jiang2017heterogeneity,2017arXiv171002368H} suggest to specify a projection function
$\Psi(\cdot)$ that modifies $\Delta\theta_t$ such that the instability that arises from stale updates is mitigated. That is,
\begin{equation}
  \label{eq:update_psi}
  {\theta}_{t+1} = {\theta}_t + \Psi(\Delta\theta_t).
\end{equation}
In particular, considering staleness to have a negative effect, $\lambda$-\textsc{softsync}~\citep{2015arXiv151105950Z} and
\textsc{dynsgd}~\citep{jiang2017heterogeneity} make use of $\tau_t$ to weigh down
an update $\Delta\theta_t$.
While there is a significant amount of
empirical evidence that these methods are able to converge, there
is also an equivalent amount of theoretical and experimental evidence that shows
that parameter staleness can actually be beneficial, especially when the number of asynchronous workers is
small~\citep{2016arXiv160509774M, 2017arXiv170603471Z,kurth2017deep}.
Most notably, \citep{2016arXiv160509774M} identifies that asynchrony induces 
an implicit update momentum, which is beneficial if kept under control, but that can otherwise have a negative effect when being too strong. 
Clearly, these results
suggest that approaches which use $\tau_t$, or the number of workers $n$,
are impairing the contribution of individual workers. Of course, this is not desired. As a result, asynchronous methods
that are commonly used in practice are variants of
\textsc{downpour}~\citep{dean2012large} or
\textsc{hogwild!}~\citep{2011arXiv1106.5730N}, where the number of asynchronous
workers is typically restricted or the learning rate lowered to ensure stability while at the same time leveraging
effects such as implicit momentum. As the number of
workers increases, one might wonder what would be an effective way
to measure how distributed asynchronous \textsc{sgd} adheres to the desired
dynamics of a stable process. Answering this question requires a framework in which we are able
to quantify this, along with a definition of stability and desired behavior.



\paragraph{Contributions}
Our contributions are as follows:
\begin{enumerate}
\item We formulate stochastic gradient descent in the context of Lagrangian mechanics
and derive a sufficient condition for ensuring the stability of a distributed asynchronous system.
\item Building upon this framework, we propose a variant of distributed asynchronous \textsc{sgd}, \textsc{gem}, that views the set of active workers as a whole and adjusts individual worker updates in order to match the dynamics of a target synchronous process.
\begin{itemize}
\item Contrary to previous methods, this paradigm scales to a large number of asynchronous workers, as convergence is guaranteed by the target dynamics.
\item Speedup is attained by partitioning the effective mini-batch size over the workers, as the system is constructed such that they collectively adhere to the target dynamics.  
\end{itemize}
\end{enumerate}

\begin{table}[htbp]\caption{Notation summary}
\begin{center}
\begin{tabular}{c p{10cm} }
\toprule
$\eta$ & Learning rate \\
$\mu$ & Momentum term \\
$t$ & Time at global (parameter server) clock. \\
${\theta}_t$ & Central variable (central parameterization) at time $t$. \\
$\tau_t$ & Staleness of worker $w$ responsible for update $t$. \\
$s_t$ & The parameterization ${\theta}_{t-\tau_t}$ of the central variable used by worker $w$ responsible for update $t$.\\
$\Delta\theta_t$ & Update produced by worker $w$ responsible for update $t$. In this work, $\Delta\theta_t$ is computed using \textsc{sgd}, i.e., $\Delta\theta_t = -\eta\nabla_\theta\mathcal{L}(s_t)$. \\
$\vert \Delta\theta_t\vert$ & Absolute value of an update $\Delta\theta_t$. \\
$\mathcal{E}_k(t+1)$ & Kinetic energy of the proxy at time $t+1$. \\
${E}_k(t+1)$ & Kinetic energy of the central variable at time $t+1$. \\
\bottomrule
\end{tabular}
\end{center}
\label{tab:TableOfNotationForMyResearch}
\end{table}


\section{Stability from energy}
\label{sec:parameter_staleness}

Taking inspiration from physics, the traditional explicit momentum \textsc{SGD}
update rule can be re-expressed within the framework of Lagrangian mechanics,
as shown in
Equations~\ref{eq:sgd_lagrangian} and~\ref{eq:sgd_dissipation}, where $D$ is
Rayleigh's dissipation function to describe the non-conservative friction force:
\begin{align}
  &L \triangleq \frac{1}{2}\dot{\theta}^2 - \eta\mathcal{L}(\theta) \label{eq:sgd_lagrangian} \\
  &D \triangleq \frac{1}{2}(\mu\dot{\theta})^2 \label{eq:sgd_dissipation}
\end{align}
\begin{equation}
  \frac{d}{dt}\left(\frac{\partial L}{\partial \dot{\theta}}\right) - \frac{\partial L}{\partial \theta} = \frac{\partial D}{\partial \dot{\theta}} \label{eq:euler_lagrange}
\end{equation}
Solving the Euler-Lagrange equation in terms of Equations~\ref{eq:sgd_lagrangian} and~\ref{eq:sgd_dissipation} yields the dynamics
\begin{equation}
  \label{eq:lagrangian_solved}
  \ddot{\theta} = \mu\dot{\theta}-\eta\nabla_\theta\mathcal{L}(\theta),
\end{equation}
which is equivalent to the traditional explicit momentum \textsc{sgd} formulation. Since the Euler-Lagrange equation can be used to derive the equations of motion using energy terms, we can similarly express these energies in an optimization setting. Using Equation~\ref{eq:sgd_lagrangian}
and discretizing the update process, we find that the kinetic energy $E_k$ and the potential energy $E_p$ are defined at time $t+1$ as
\begin{align}
  E_k(t+1) &= \frac{1}{2}\dot{\theta}_{t+1}^2 = \frac{1}{2}\Big(\theta_{t+1} - \theta_{t}\Big)^2 \label{eq:kinetic_energy} \\
  E_p(t+1) &= -\eta\mathcal{L}(\theta_{t+1}) \label{eq:potential_energy}
\end{align}
where the loss function $\mathcal{L}$ describes the potential under a learning rate $\eta$.

Equations~\ref{eq:kinetic_energy} and~\ref{eq:potential_energy} only
hold in a sequential (or synchronous) setting as the
velocity term $\dot{\theta}_{t+1}$ is ill-defined in the asynchronous case. 
However,
since asynchronous methods are concerned with the optimization of the central
variable, a worker $w$ can implicitly observe the velocity of ${\theta}_{t+1}$, that is the parameter shift, 
between the moment it read the central variable and the moment it committed its update. In the asynchronous setting,
\begin{equation}
    {\theta}_{t+1} = {\theta}_{t} + \Delta\theta_{t}
\end{equation}
where $\Delta\theta_{t}$ was computed from $s_{t} = {\theta}_{t - \tau_{t}}$. Therefore,
the parameter shift between the time worker $w$ responsible for the update $t$ read $s_{t}$ and eventually committed $\Delta\theta_{t}$ is ${\theta}_{t+1} - {\theta}_{t - \tau_{t}}$.
From this, we define the kinetic energy ${E}_k(t+1)$ of the central variable as observed from $w$ to produce ${\theta}_{t+1}$ as
\begin{equation}
    \label{eq:generalized_energy}
    {E}_k(t+1) = \frac{1}{2}\Big({\theta}_{t+1} - {\theta}_{t - \tau_{t}}\Big)^2  = \frac{1}{2}\Big(\sum_{i=0}^{\tau_{t}} \Delta\theta_{t-i} \Big)^2.
\end{equation}
Intuitively, our definition of the kinetic energy of the asynchronous system therefore corresponds to the collective kinetic energy induced by the last updates from every worker responsible for $\tau_t$. 

This definition allows us to measure the kinetic energy of the central variable. From this, we may now formulate a sufficient
condition ensuring the stability and compliance of the asynchronous
system. To accomplish this, we introduce the kinetic energy of a proxy,
$\mathcal{E}_k(t+1)$, for which we know, by assumption, that it
eventually converges to a stationary solution $\theta^*$. As a result, an
asynchronous system can be said to be stable whenever
\begin{equation}
    \label{eq:stability}
    {E}_k(t+1) \leq \mathcal{E}_k(t+1),
\end{equation}
and compliant when
\begin{equation}
    \label{eq:stability_preferred}
    {E}_k(t+1) = \mathcal{E}_k(t+1).
\end{equation}
That is, stability and compliance are ensured whenever the kinetic energy collectively induced by the set of workers matches with the kinetic energy of a proxy known to converge.
Therefore, equations~\ref{eq:stability}
and~\ref{eq:stability_preferred} are sufficient to ensure convergence, albeit not necessary. 


\section{Gradient Energy Matching}
\label{sec:gradient_energy_matching}

We now
present a new variant of distributed asynchronous \textsc{sgd}, Gradient Energy Matching (\textsc{gem}),
which aims at satisfying the compliance condition with respect to the dynamics of a target synchronous process, thereby ensuring its stability. In addition, we design \textsc{gem} so as to minimize computation at the master node, thereby reducing the risk of bottleneck.

\subsection{Proxy}
\label{sec:target_proxy}

The definition of the proxy can be chosen arbitrarily, as it depends on the desired dynamics of the central variable.
In this work, we target a proxy for which the dynamics of ${\theta}_t$ behaves similarly to what would be computed sequentially in \textsc{sgd} with an explicit momentum term $\mu$ under a static learning rate $\eta$. Traditionally, this is written as
\begin{equation}
    \label{eq:momentum_update}
    \Delta\theta_{t} = \mu\Delta\theta_{t-1} - \eta\nabla_\theta\mathcal{L}(\theta_{t}),
\end{equation}
with the corresponding kinetic energy
\begin{equation}
    \label{eq:proxy}
    \mathcal{E}_k(t+1) \triangleq \frac{1}{2}\left(\mu\Delta\theta_{t-1} - \eta\nabla_\theta\mathcal{L}(\theta_{t})\right)^2.
\end{equation}
\subsection{Energy matching}
\label{sec:imperfect_strategies}
Consider the situation of a small number of asynchronous workers, as is
typically  the case for \textsc{downpour} or \textsc{hogwild!}. Experimental results reported in the literature suggest that these methods usually converge to a
stationary solution, which implies that the stability condition
(Equation~\ref{eq:stability}) is most likely satisfied. However, as the number of
asynchronous workers increases, so does the kinetic energy of the central
variable and there is a limit beyond which the system dynamics do no
longer satisfy Equation~\ref{eq:stability}. To address this, we introduce
scaling factors $\pi_{t}$, whose purpose are to modify updates $\Delta\theta_{t}$ such
that the compliance condition remains satisfied. As a result, the central
variable energy (Equation~\ref{eq:generalized_energy}) 
is extended to
\begin{equation}
    \label{eq:central_variable_energy_implicit}
    E_k(t + 1) \triangleq \frac{1}{2}(\theta_t - s_t + \pi_t\Delta\theta_t)^2,
\end{equation}
Scaling factors $\pi_{t}$ can be assigned in several ways, depending on the information available to each worker $w$ and on the desired usage of the individual updates $\Delta\theta_{t}$. To prevent wasting computational resources, we aim at equally maximizing the effective contribution of all updates $\Delta\theta_{t}$ while achieving compliance with the proxy, i.e.,
\begin{align}
    \label{eq:proxy_compliance}
    \mathcal{E}_k(t + 1) &= E_k(t + 1), \\
    &= \frac{1}{2}(\theta_t - s_t + \pi_t\Delta\theta_t)^2.
\end{align}
Directly solving for $\pi_t$ yields
\begin{equation}
    \label{eq:new_pi}
    \pi_t = \frac{\sqrt{2}\sqrt{\mathcal{E}_k(t+1)} - \vert\theta_t-s_t\vert}{\vert\Delta\theta_t\vert}.
\end{equation}
This mechanism constitutes the core of \textsc{gem}, as it is through scaling factors $\pi_t$ that the energy of the active workers is matched to the energy of the target proxy. 

To show the reader how $\pi_t$ adjusts the active workers to collectively adhere to the proxy, let us consider the quantity $\vert\theta_t - s_t\vert$. From~\cite{2016arXiv160509774M}, we know that under a homogeneity assumption the average staleness is $\mathbb{E}_\mathcal{Q}[\tau] = n - 1$. In expectation, the behavior of the system is therefore similar to updates occurring in a round-robin fashion. In this case, Equation~\ref{eq:generalized_energy} reduces to
\begin{align}
    \label{eq:central_variable_energy_ideal}
    {E}_k(t+1) &\triangleq \frac{1}{2}\left(\sum_{i=0}^{n-1} \Delta\pi_{t - i}\theta_{t-i}\right)^2, \\
    &\equiv \frac{1}{2}\left(\pi_t\Delta\theta_t + \sum_{i=1}^{n-1} \pi_{t - i}\Delta\theta_{t-i}\right)^2, \\
    &\equiv \frac{1}{2}\left(\pi_t\Delta\theta_t + \vert\theta_t - s_t\vert\right)^2,
\end{align}
which shows that to ensure compliance, the updates of all workers have to collectively be scaled. 


\begin{algorithm}[t]
  \caption{Worker procedure of Gradient Energy Matching.}
  \emph{Hyper-parameters}: learning rate $\eta = 0.01$, momentum $\mu = 0.9$, number of asynchronous workers $n$, and $\epsilon = 10^{-16}$ to prevent division by zero.
  \newline
  \newline
  \label{algo:gem_imperfect}
  \begin{minipage}[t]{.5\linewidth}
    \begin{algorithmic}[1]
      \Procedure{$\Psi_\textsc{gem}$}{$\Delta\theta$}
        \State $m \gets \mu m + \Delta\theta$
        \State $\pi \gets \frac{\vert m\vert - \vert\theta - s\vert}{\vert\Delta\theta_t\vert + \epsilon}$
        \State \textbf{return} $\pi \odot \Delta\theta$
      \EndProcedure
    \end{algorithmic}
  \end{minipage}
  \begin{minipage}[t]{.5\linewidth}
    \begin{algorithmic}[1]
      \Procedure{Worker}{}
      \State $m \gets 0$
      \State $s \gets \Call{Pull}{~}$
        \While{\textbf{not} \emph{converged}}
          \State $\Delta\theta \gets -\eta\nabla_\theta\mathcal{L}(s)$
          \State $\theta \gets \Call{Pull}{~}$
          \State \Call{Commit}{{$\Psi_\textsc{gem}(\Delta\theta)$}}
          \State $s \gets \theta$
        \EndWhile
      \EndProcedure
    \end{algorithmic}
  \end{minipage}
\end{algorithm}


\subsection{Proxy estimation}
\label{sec:proxy_estimation}

While the previous derivation defines an assignment for $\pi_{t}$, it remains to locally estimate the desired kinetic energy $\mathcal{E}_k(t+1)$. This is particularly problematic since updates $\Delta\theta_t$ from the other workers are not available. As an approximation, we propose to aggregate local updates such that
\begin{equation}
    \label{eq:proxy_estimate}
    \mathcal{E}_k(t+1) \triangleq \frac{1}{2}\left(\mu m_{t-\tau_{t}} - \eta\nabla_\theta\mathcal{L}(s_{t})\right)^2,
\end{equation}
with the first moment $m_{t}$ being defined as
\begin{equation}
    \label{eq:moment_estimate}
    m_t \triangleq \mu m_{t-\tau_t} - \eta\nabla_\theta\mathcal{L}(s_t).
\end{equation}
Additionally, Appendix~\ref{sec:proxy_definitions} examines the effects of scaling $\mathcal{E}_k(t+1)$ with an amplification factor to compensate for the staleness of the proxy due to the local approximation.

This concludes \textsc{gem}, as summarized in Algorithm~\ref{algo:gem_imperfect}.

\section{Experiments}
\label{sec:experiments}

We evaluate \textsc{gem} against \textsc{downpour}~\cite{dean2012large} and an adaptive staleness technique which rescales updates $\Delta\theta_t$ by $1 / (\tau_t+1)$, similar to what is typically applied to stabilize stale gradients in $\lambda$-\textsc{softsync}~\citep{2015arXiv151105950Z} or
\textsc{dynsgd}~\citep{jiang2017heterogeneity}. We carry out experiments on MNIST~\citep{mnist} and ImageNet~\citep{imagenet} to illustrate the
effectiveness and robustness of our method. All algorithms are implemented in
PyTorch~\citep{paszke2017automatic} using the distributed sub-module. The
source code~\footnote{\url{https://github.com/montefiore-ai/gradient-energy-matching}} we provide is to the best of our knowledge the first PyTorch-native
parameter server implementation. Experiments described in this section are carried out on general purpose CPU machines without using GPGPU accelerators.

\subsection{Illustrative experiments}
\label{sec:experiment_handwritten_digits}

\paragraph{Setup} For all experiments described below we train a convolutional network (see Appendix~\ref{sec:mnist_model} for specifications) on MNIST using \textsc{gem}, \textsc{downpour} or the adaptive staleness variant on a computer cluster as an MPI job. A fixed learning rate $\eta = 0.05$, momentum term $\mu = 0.9$ and worker batch-size $b = 64$ are applied unless stated otherwise. 
Additionally, every worker contributes a single epoch of training data. All results in this section is the average of 3 consecutive runs.

\begin{figure}
    \centering
    \begin{subfigure}[b]{0.49\textwidth}
        \includegraphics[width=\textwidth]{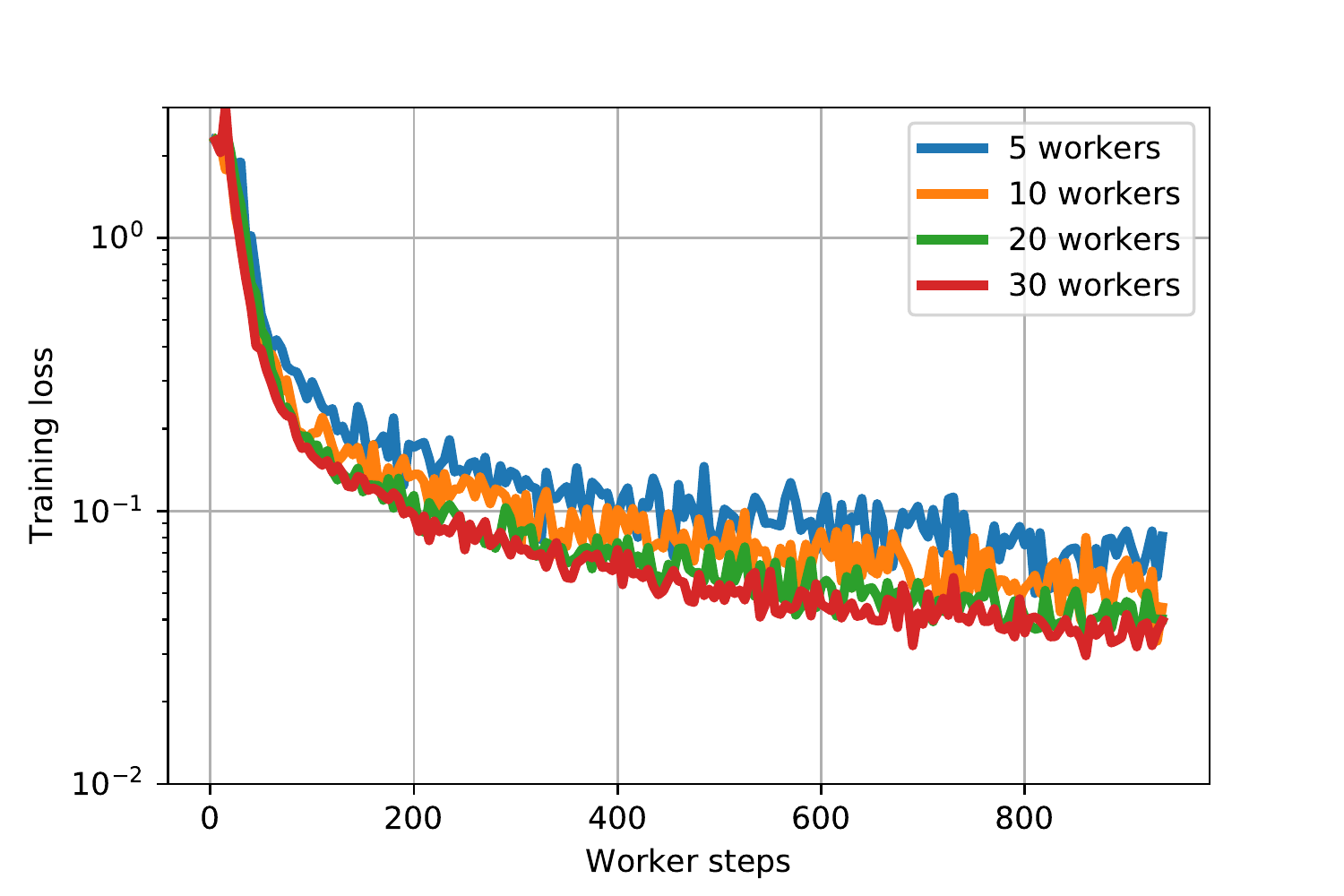}
        \caption{\textsc{gem}}
        \label{fig:comparison-gem}
    \end{subfigure}
    \begin{subfigure}[b]{0.49\textwidth}
        \includegraphics[width=\textwidth]{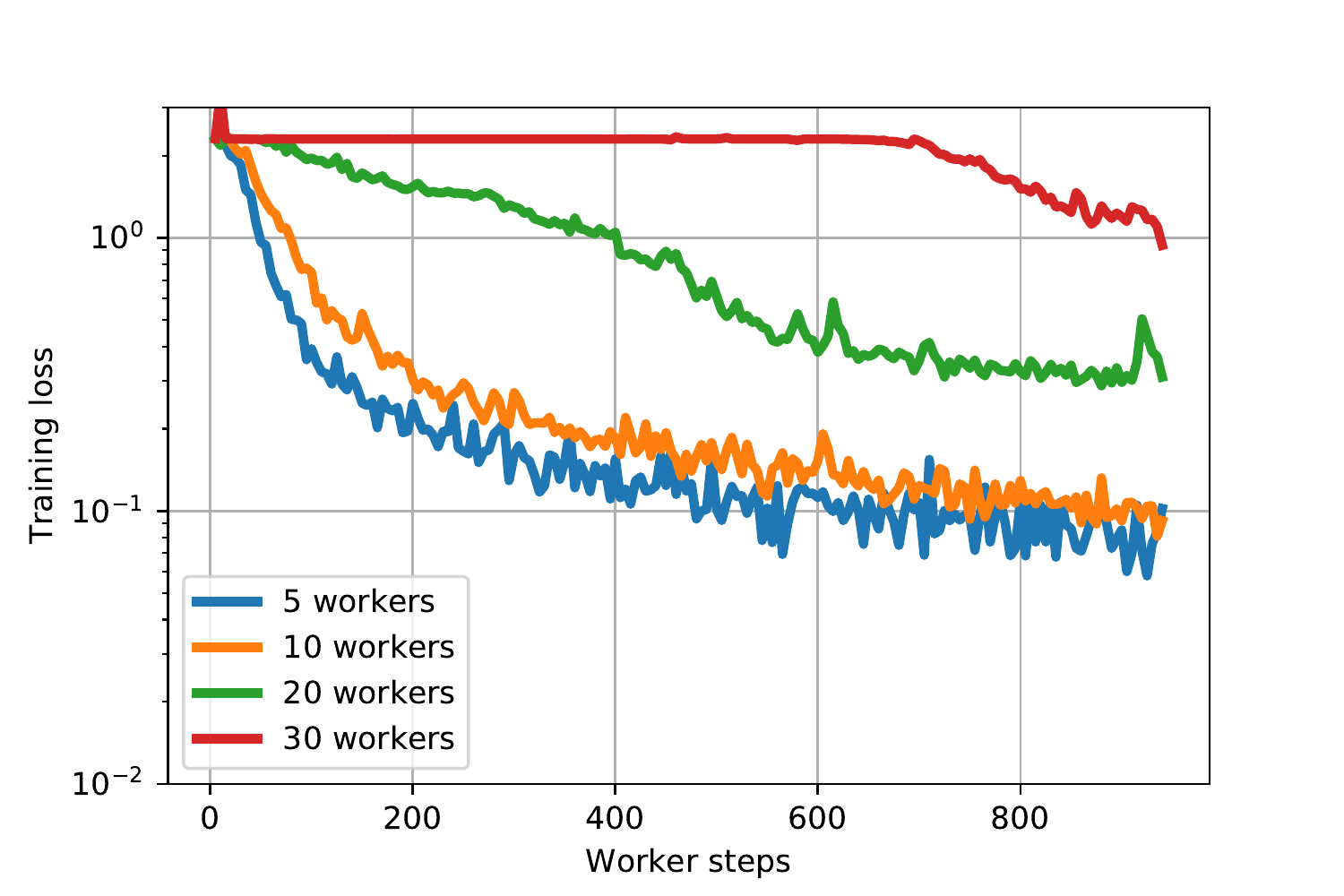}
        \caption{\textsc{downpour}}
        \label{fig:comparison-downpour}
    \end{subfigure}
    \begin{subfigure}[b]{0.49\textwidth}
        \includegraphics[width=\textwidth]{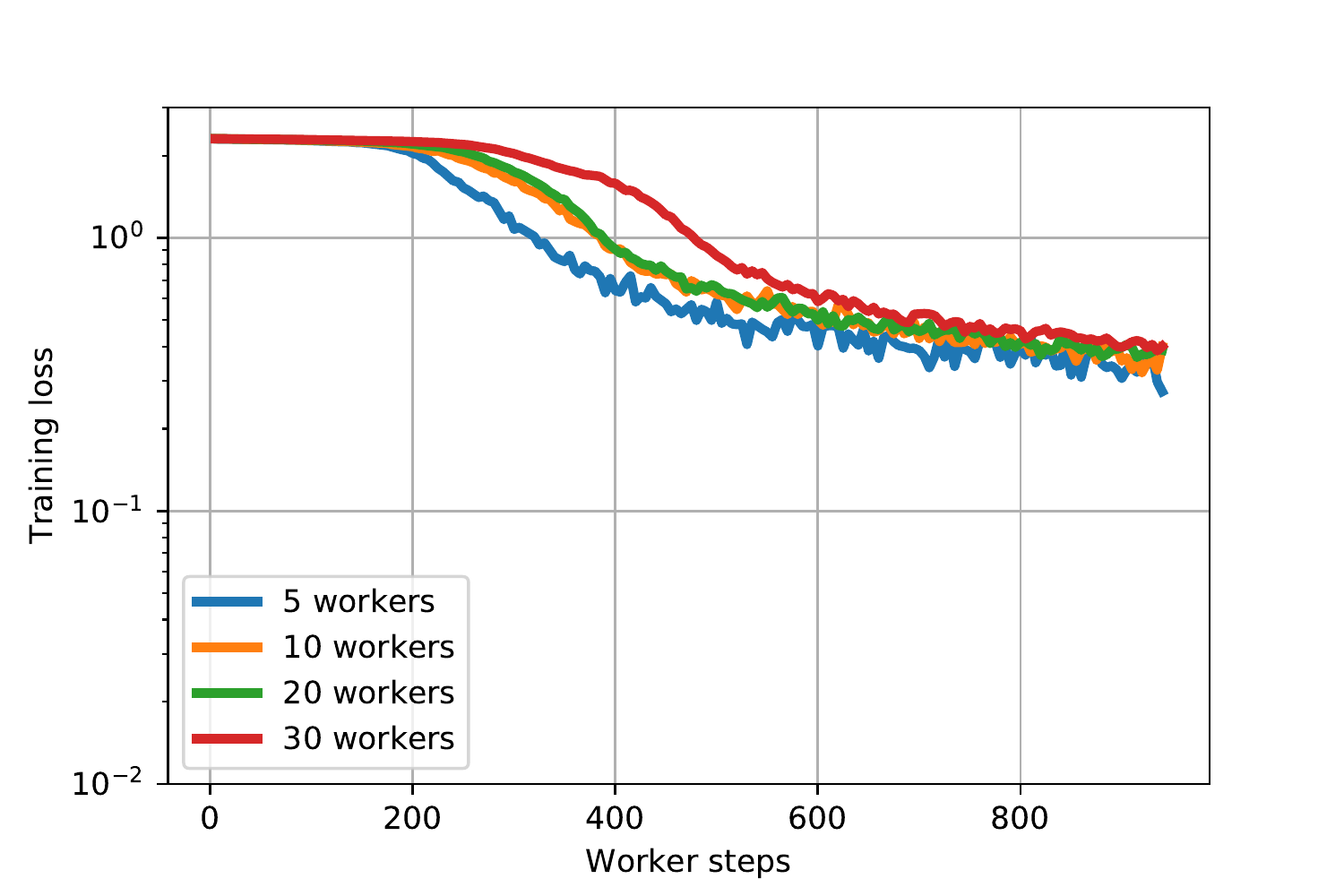}
        \caption{Adaptive staleness}
        \label{fig:comparison-as}
    \end{subfigure}
    \begin{subfigure}[b]{0.49\textwidth}
        \includegraphics[width=\textwidth]{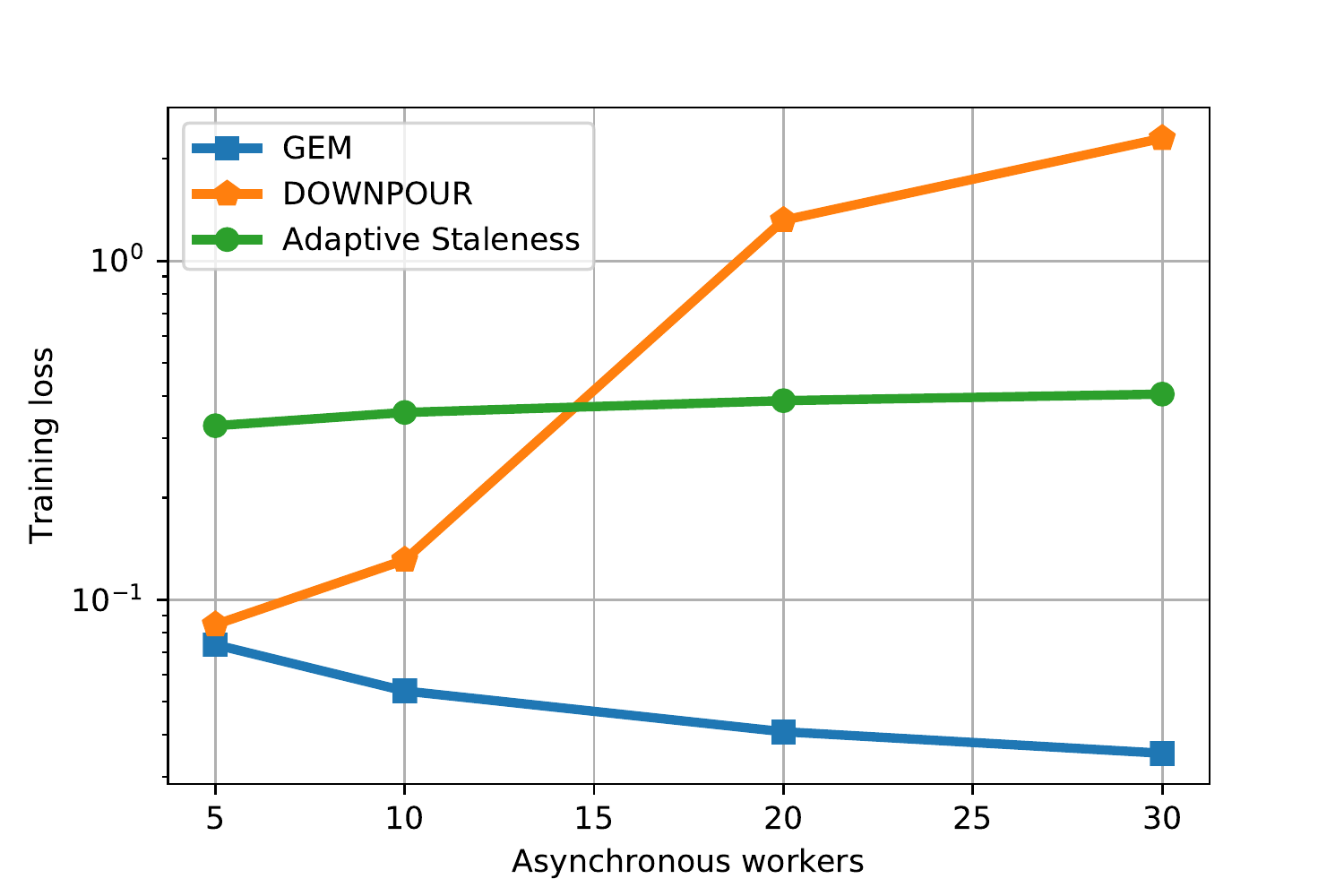}
        \caption{Summary}
        \label{fig:comparison-summary}
    \end{subfigure}
    \caption{(a) Training loss of \textsc{gem} for different number of workers. We observe that the optimizer adheres to the specified target dynamics, and that the variance of the loss becomes smaller when the number of workers increases. (b) \textsc{downpour} on the other hand quickly shows suboptimal or even divergent behavior when the number of workers is increased. (c) The adaptive staleness technique is slower since the updates are scaled  proportionally to their staleness $\tau$. (d) Final achieved training loss given the number of workers, which highlights the stability of \textsc{gem}.}
    \label{fig:comparison}
\end{figure}

\begin{figure}
    \centering
    \begin{minipage}[t]{.5\textwidth}
        \centering
        \captionsetup{width=.95\linewidth}
        \includegraphics[width=0.98\textwidth]{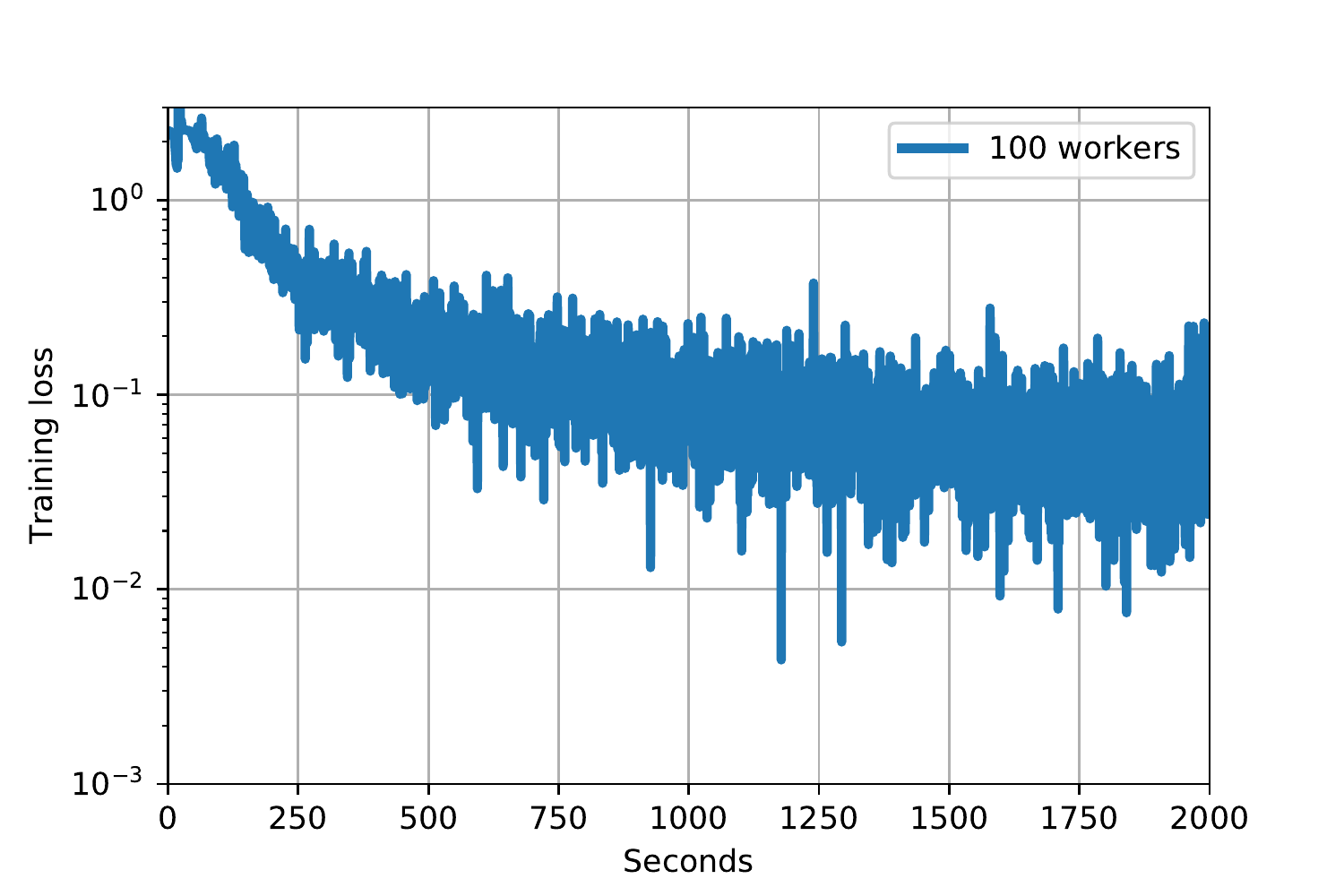}
        \caption{Training loss at the master node for 100 asynchronous workers. Even at that scale, \textsc{gem} does not show signs of instability.}
        \label{fig:gen}
    \end{minipage}\begin{minipage}[t]{0.5\textwidth}
        \centering
        \captionsetup{width=.95\linewidth}
        \includegraphics[width=0.98\textwidth]{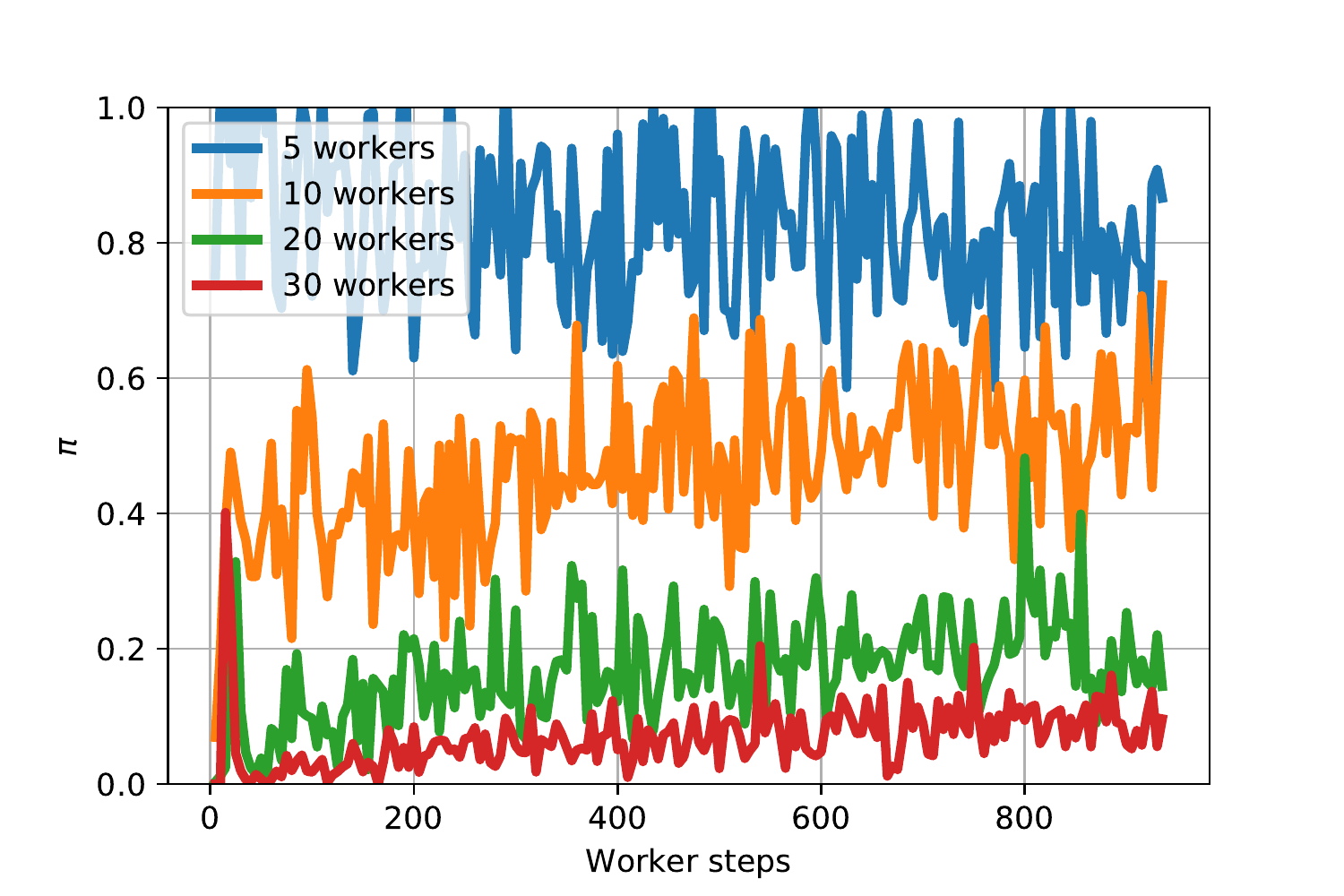}
        \caption{Change over time of the median value of the scaling factors $\pi$.}
        \label{fig:scalability}
    \end{minipage}
    \begin{minipage}[t]{.5\textwidth}
        \centering
        \captionsetup{width=.95\linewidth}
        \includegraphics[width=0.98\textwidth]{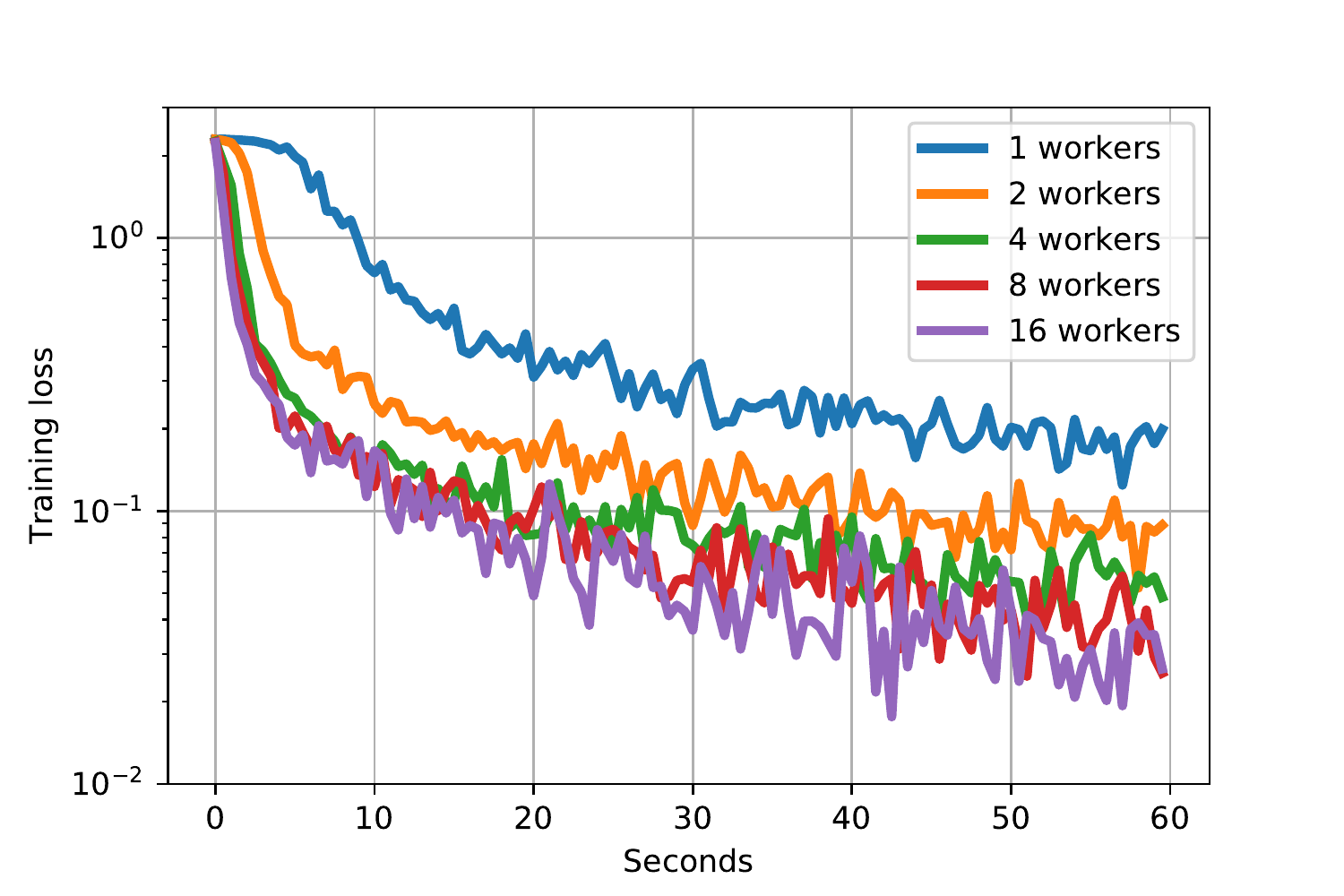}
        \caption{In \textsc{gem}, wall-clock speedup is obtained by splitting the effective batch-size over $n$ workers, as is done in synchronous methods.}
        \label{fig:speedup}
    \end{minipage}\begin{minipage}[t]{0.5\textwidth}
        \centering
        \captionsetup{width=.95\linewidth}
        \includegraphics[width=0.98\textwidth]{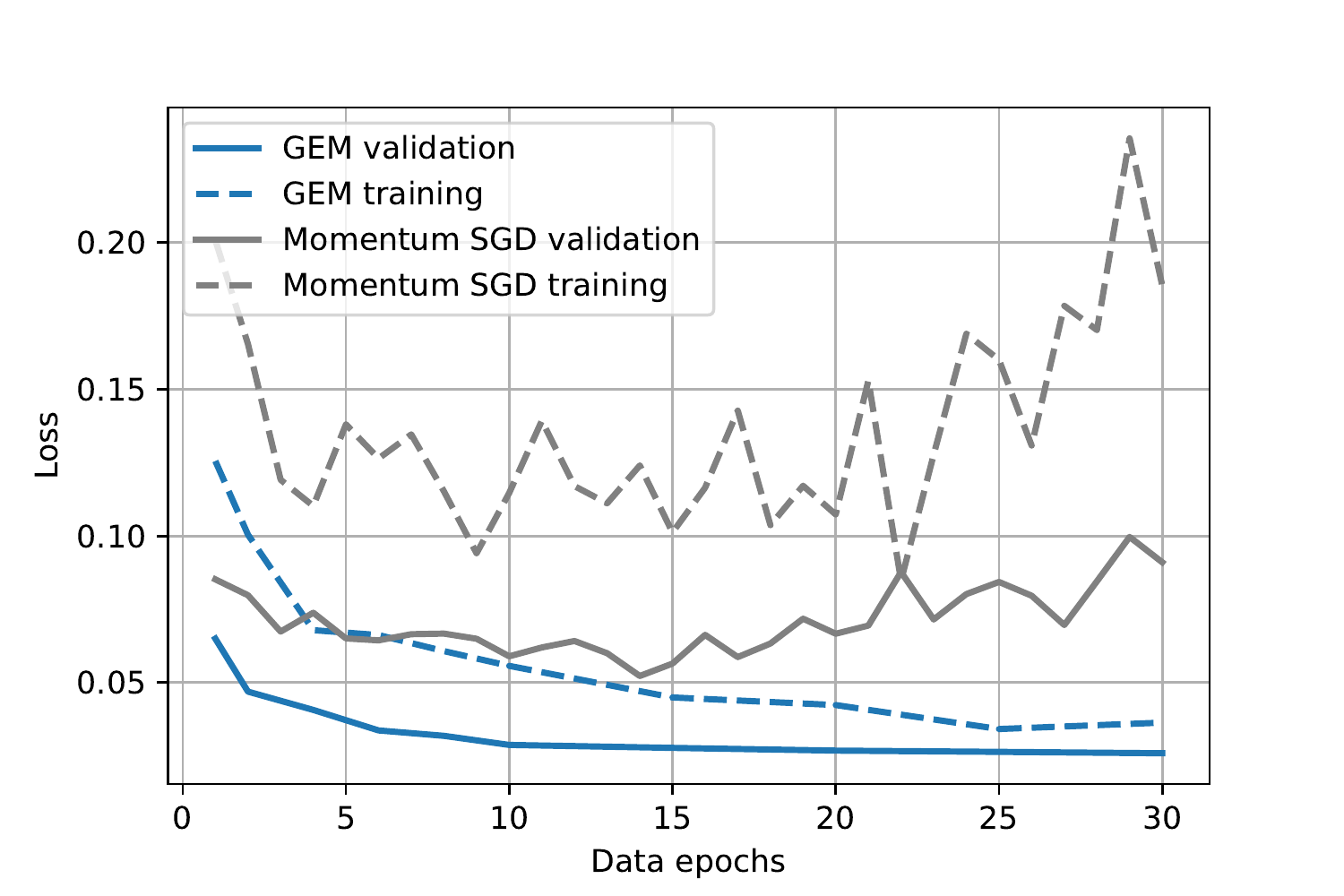}
        \caption{Training and validation losses for \textsc{gem} and the proxy. Contrary to momentum \textsc{sgd}, \textsc{gem} does not show signs of overfitting.}
        \label{fig:overfitting}
    \end{minipage}
\end{figure}

\paragraph{Stability} Figure~\ref{fig:comparison} illustrates the training loss of \textsc{gem} compared to \textsc{downpour} and adaptive staleness. We observe that \textsc{gem} remains stable and adheres to the target dynamics as designed, even when increasing the number of asynchronous workers. This shows that \textsc{gem} is able to cope with the induced staleness.
Remarkably, \textsc{gem} remains stable even when scaling to 100 asynchronous workers, as shown in Figure~\ref{fig:gen}.
By contrast, \textsc{downpour} faces strong convergence issues when increasing the number of workers while adaptive staleness shows impaired efficiency as updates are rescaled proportionally to the staleness, which is often too aggressive. 

\paragraph{Update rescaling} Figure~\ref{fig:comparison-gem} shows the behavior of \textsc{gem} for an increasing number of workers and a fixed worker batch-size $b$. When applying an asynchronous method, one would expect that the  number of updates per worker to reach a given loss value would decrease since more updates are committed in total to the master node. However, since \textsc{gem} tries to make the workers collectively adhere to the dynamics that are expressed by the proxy, this behavior should not be expected. Indeed, Figure~\ref{fig:comparison-gem} shows loss profiles that are similar whereas the number of workers varies from 5 to 30. As designed, inspecting the median per-parameter value of $\pi$, shown in Figure~\ref{fig:scalability}, reveals that the scaling factors steadily decrease whenever additional workers are added. 
Notwithstanding, we also do observe improvement as we increase the number of workers as it indirectly augments the effective batch size $B$. 
Furthermore, we observe from Figure~\ref{fig:scalability} the adaptivity of \textsc{gem} as the median $\pi$ changes with time. While this behavior may appear as a limitation, better efficiency can be achieved by simply considering proxies of different energies, as further discussed in Appendix~\ref{sec:proxy_definitions}. 

\paragraph{Wall-clock time speedup} In distributed synchronous optimization, wall-clock speedup is typically achieved by setting a desired effective batch-size $B$, and splitting the computation of the batch over $n$ workers to produce updates $\Delta\theta_t$ computed individually over $b=B/n$ samples. In the asynchronous setting, the same strategy can be applied for \textsc{gem} to speed up the training procedure,  as shown in Figure~\ref{fig:speedup} for a fixed effective batch-size $B = 256$, a number of workers $n$ between 1 and 16 and a corresponding worker batch size $b$ between 256 and 16. 
As expected, and without loss in accuracy, we observe that wall-clock speedup is inversely proportional to $b$, depending on the efficiency of the implementation. 
However, we also observe larger variability as $n$ increases due to the increased stochasticity of the updates.   

\paragraph{Generalization} A common argument against large mini-batch \textsc{sgd} is the poor generalization performance that is typically associated with the resulting model due to easily obtainable sharp minima~\cite{keskar2016large, hochreiter1997flat, hochreiter1995simplifying, chaudhari2016entropy}. Nevertheless, \cite{smith2017don, goyal2017accurate, krizhevsky2014one} show that test accuracy is not significantly impacted when using larger learning rates to emulate the regularizing effects of smaller batches. Similarly, experimental results for \textsc{gem} do not show such behavior. Even when training with 100 asynchronous workers, as done in Figure~\ref{fig:gen} and which corresponds to $B=nb=100\times 64$, \textsc{gem} obtains a final test accuracy of 99.38\%. By contrast, the targeted momentum \textsc{sgd} procedure starts to overfit, as shown in Figure~\ref{fig:overfitting}. A possible intuitive explanation for this behavior is most in-line with the arguments presented in~\cite{chaudhari2016entropy, keskar2016large}, i.e., that asynchrony (staleness) regularizes the central variable such that it is not attracted to sharper minima, therefore preferring wider basins.


\subsection{Resilience against hyper-parameter misspecification}
\label{sec:hyperparam_misspecification}

\begin{figure}
    \centering
    \begin{subfigure}[b]{0.49\textwidth}
            \includegraphics[width=\textwidth]{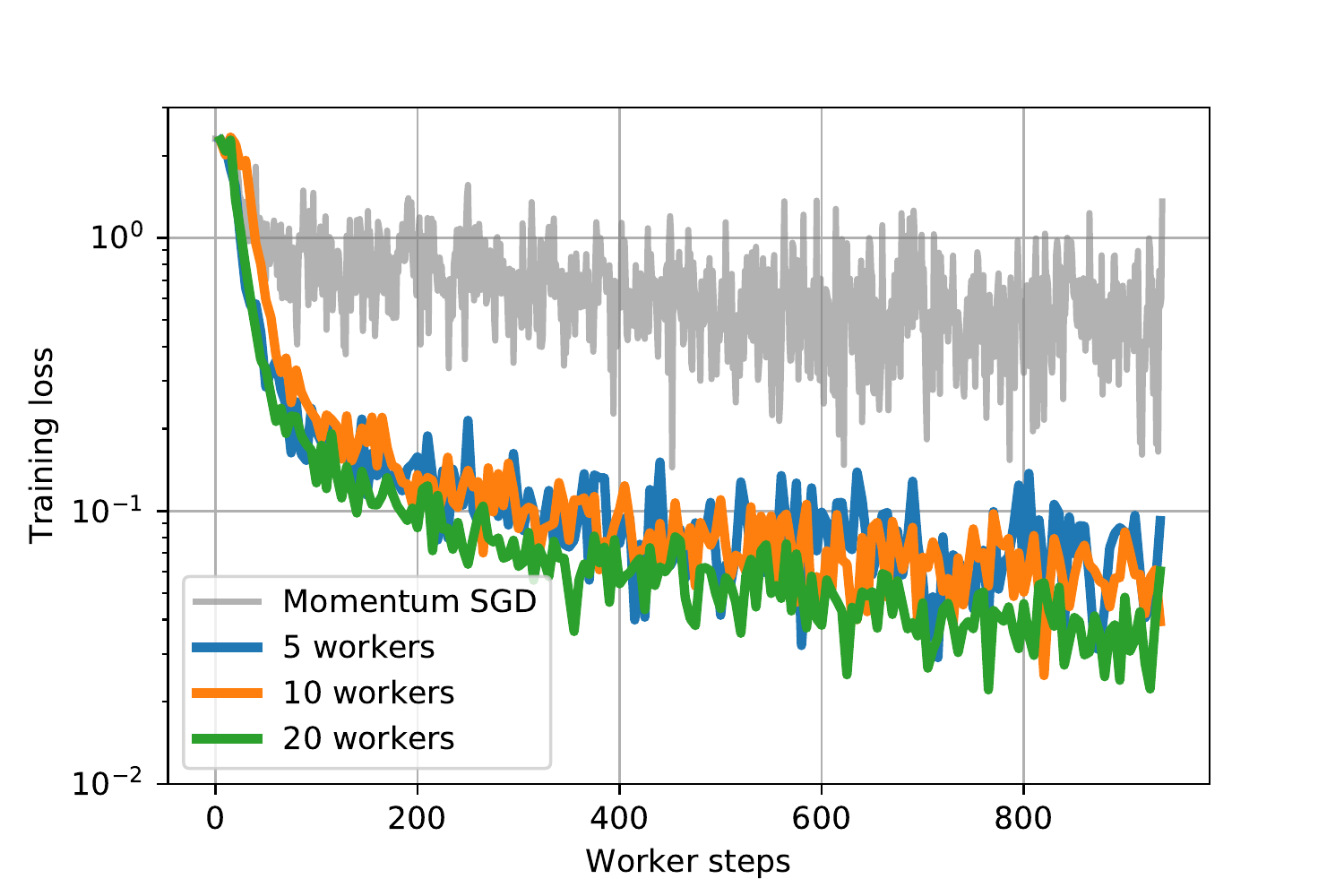}
            \caption{\textsc{gem}}
    \end{subfigure}
    \begin{subfigure}[b]{0.49\textwidth}
            \includegraphics[width=\textwidth]{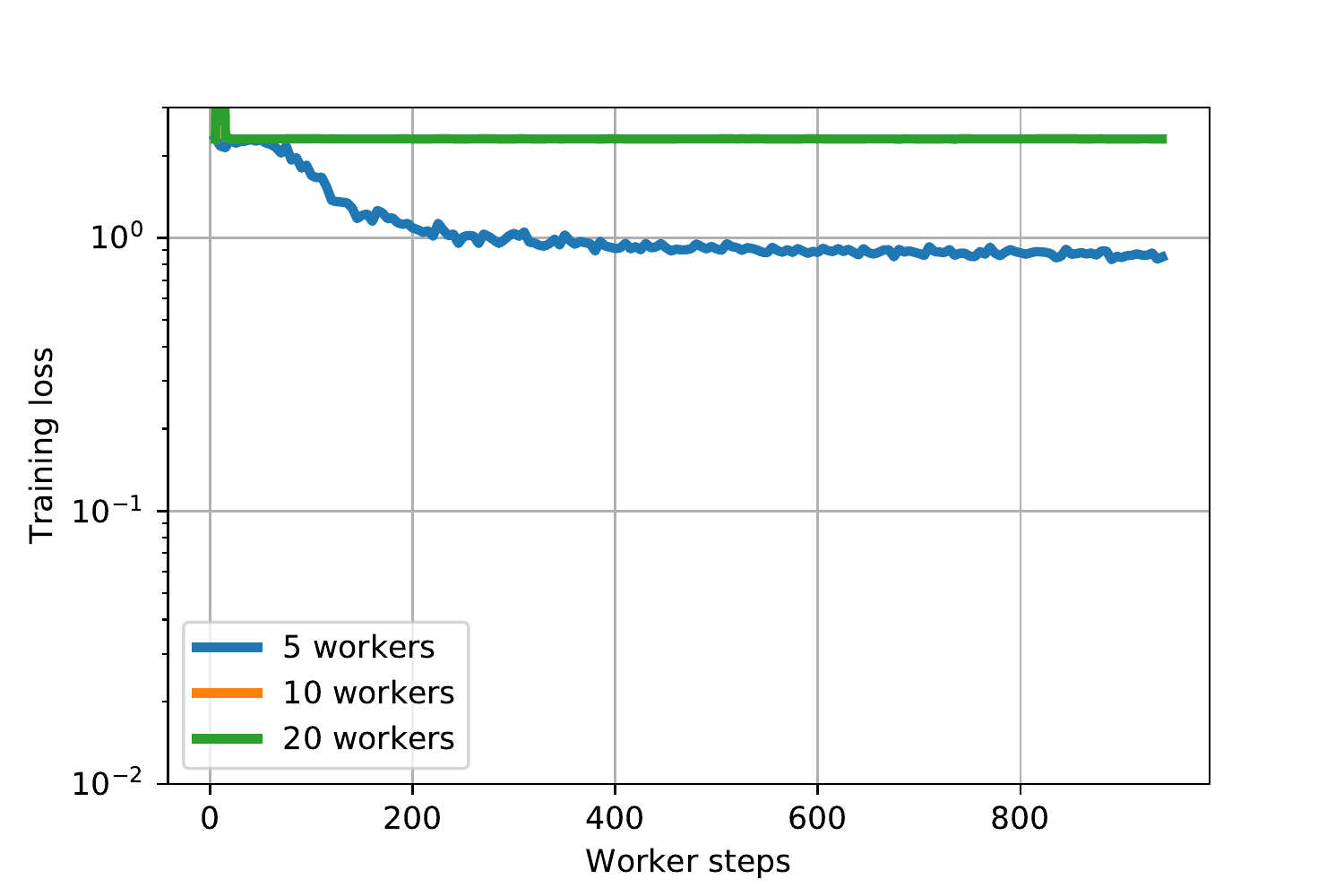}
            \caption{\textsc{downpour}}
    \end{subfigure}
    \caption{Training curves for \textsc{gem} and \textsc{downpour} when the learning rate is misspecified  to $\eta = 0.1$. \textsc{downpour} clearly shows suboptimal behavior while \textsc{gem} and the target synchronous momentum \textsc{sgd} proxy converge without issues.}
    \label{fig:misspecification}\vspace{-1em}
\end{figure}

These experiments use the same setup as in the previous MNIST experiments, with the exception of the learning rate, which is intentionally set  to a large value $\eta = 0.1$.
To show that our method is able to adhere to the proxy dynamics, we compare \textsc{gem} against \textsc{downpour}. From Figure~\ref{fig:misspecification}, we observe that \textsc{downpour} is not able to cope with the increased learning rate, while \textsc{gem} remains stable since the equivalent synchronous process converges as well under these settings.

\subsection{ImageNet}
\label{sec:experiment_imagenet}

\begin{figure}
    \centering
    \includegraphics[width=0.49\textwidth]{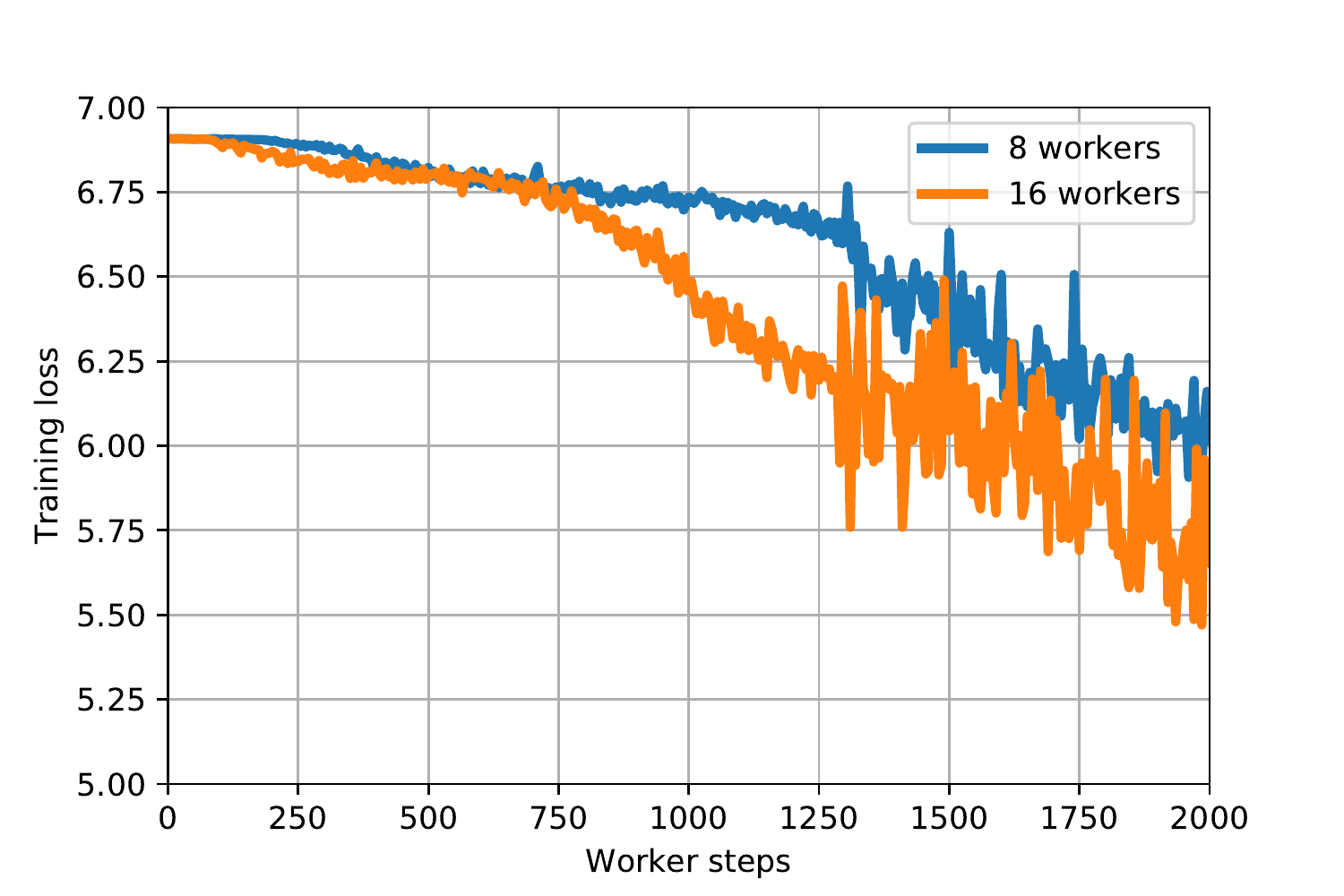}
    \caption{ImageNet experiments. All runs have a fixed time-budget of 24 hours. Training loss with respect to the worker steps. We observe a significant improvement in update efficiency when doubling the number of workers, while halving $b$.}
    \label{fig:imagenet_results}
\end{figure}

We train AlexNet~\cite{krizhevsky2012imagenet} on the ImageNet data~\citep{imagenet} with an effective batch size of $B = 1024$ and $b = B / n$. \textsc{gem} hyper-parameters are set to their default values, as specified in Algorithm~\ref{algo:gem_imperfect}. No learning rate scheduling is applied. All experiments ran for 24 hours with 8 and 16 asynchronous workers, each using 2 CPU cores due to cluster restrictions. A single core computes the gradients, while the other handles data-prefetching, data-processing, and logging. The results are summarized in Figure~\ref{fig:imagenet_results}. The figure shows that \textsc{gem} is able to effectively handle the staleness on a realistic use-case. Despite an identical effective batch size, we observe a significant improvement with regards to worker update efficiency, both in wall-clock time and the training loss. The reason for this effect remains to be studied.
Figures~\ref{fig:alexnet_pi_8} and \ref{fig:alexnet_pi_16} in supplementary materials show how individual tensors of an update are adjusted over time.




\section{Related work}
\label{sec:related_work}

Many techniques~\cite{schaul2013adaptive,jiang2017heterogeneity,2015arXiv151105950Z,chilimbi2014project,2017arXiv171002368H,mnih2016asynchronous,2016arXiv160509774M} have been proposed to stabilize distributed asynchronous \textsc{sgd} based on the perception that staleness usually has a negative effect on the optimization process.
In~\cite{2016arXiv160509774M} however, the main inspiration behind this work, staleness is shown to be equivalent to an asynchrony-induced implicit momentum, which thereby explains its negative effect when being too strong but also highlights its potential benefits when kept under control. 
Similarly, in our framework, implicit momentum translates to added kinetic energy of the central variable. 
Close to our work, the authors of \textsc{YellowFin}~\cite{2017arXiv170603471Z} build on top of~\cite{2016arXiv160509774M} by proposing an adaptive scheme that tunes momentum dynamically. They show in an asynchronous setting how individual workers can adjust momentum terms to collectively fit a target momentum provided by the optimizer. In particular, this is achieved by incorporating a negative feedback loop that attempts to estimate the momentum term $\mu$ from the observed staleness. We note the opportunity to extend this strategy in our framework, by using \textsc{YellowFin} to estimate the optimal momentum term of the proxy, and have \textsc{gem} acting as a mechanism to tune individual workers, thereby removing the original feed-back loop. 



\section{Conclusions}
\label{sec:conclusion_and_discussion}

This work introduces Gradient Energy Matching, \textsc{gem}, a novel algorithm for distributed asynchronous \textsc{sgd}.
The core principle of \textsc{gem} is to consider the set of active workers as a whole and to make them collectively
adhere to the dynamics of a target synchronous process.
More specifically, the estimated energy of the asynchronous system is matched with the energy of the target proxy by rescaling the gradient updates before sending them to the master node. 
Under the assumption that this proxy converges, the proposed method ensures the stability of the collective asynchronous system, even when scaling to a high number of workers. 
Wall-clock time speedup is attained by partitioning the effective mini-batch size across workers, while maintaining its stability and generalizability.

A direction for future theoretical work would be the study of the effects of asynchrony on the generalization performance in connection with~\cite{dinhsharp} and how \textsc{gem} differs from regular momentum. This could help explain the unexpected improvement observed in the ImageNet experiments. A further addition to our method would be to adjust the way the proxy is currently estimated, i.e., one could compute the proxy globally instead of locally. This would allow all workers to contribute information to produce better estimates of the (true) proxy. As a result, the scaling factors that would be applied to all worker updates would in turn be more accurate.



\subsubsection*{Acknowledgements}

Joeri Hermans acknowledges the financial support from F.R.S-FNRS for his FRIA PhD scholarship.
We are also grateful to David Colignon for his technical help.


\medskip

\bibliographystyle{plainnat}
\bibliography{biblio}


\newpage
\appendix

\section{Model specification}
\label{sec:experiment_models}
\label{sec:mnist_model}

The convolutional network in the MNIST experiments is fairly straightforward. It consists of 2 convolutional layers, each having a kernel size of 5 with no padding, with stride and dilation set to 1. The first convolutional layer has 1 input channel, and 10 output channels, whereas the second convolutional layer has 20 output channels, both with the appropriate max-pooling operations. We apply dropout~\cite{dropout} using the default settings provided by PyTorch~\cite{paszke2017automatic} on the final convolutional layer. Finally, we flatten the filters, and pass them through 3 fully connected layers with ReLU activations followed by a logarithmic softmax.

\section{High-energy proxies}
\label{sec:proxy_definitions}

\begin{figure}[h]
    \centering
    \begin{subfigure}[b]{0.49\textwidth}
        \includegraphics[width=\textwidth]{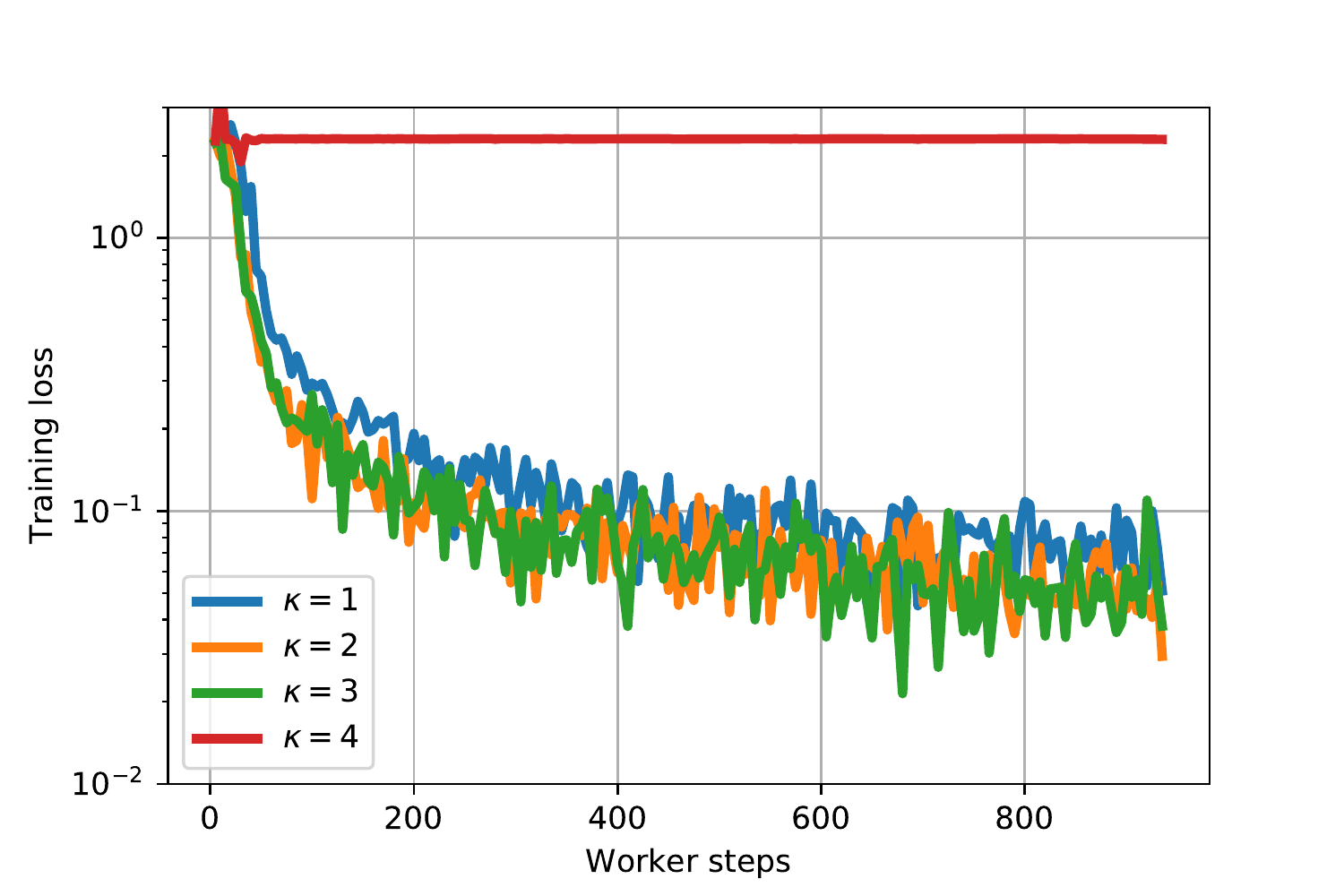}
        \caption{Training loss}
    \end{subfigure}
    \begin{subfigure}[b]{0.49\textwidth}
        \includegraphics[width=\textwidth]{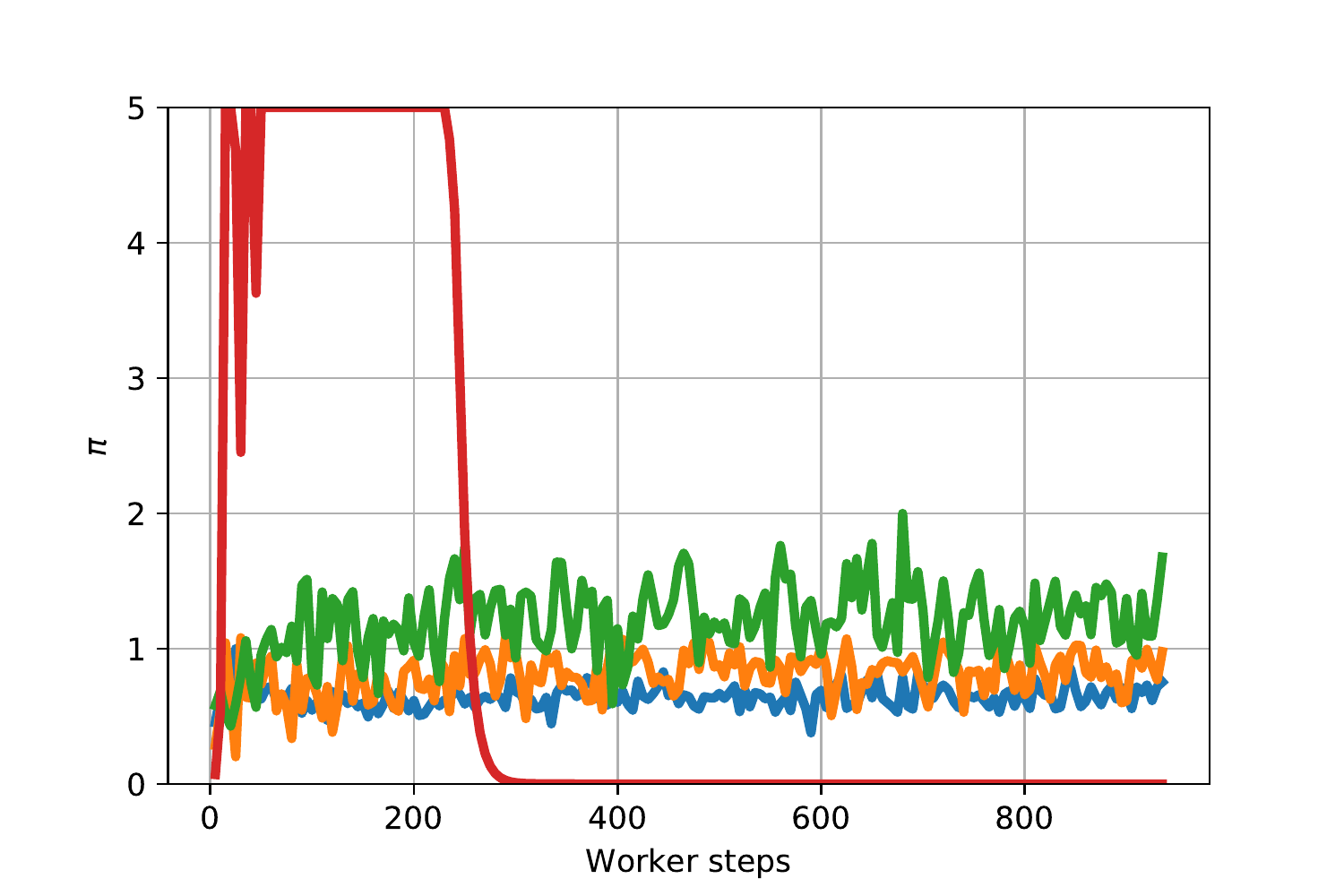}
        \caption{Median $\pi$}
    \end{subfigure}
    \caption{Effect of $\kappa$ on the convergence of rate the central variable in \textsc{gem}. Increasing the energy shows better convergence rates. However, there exists a certain limit as discussed earlier in Section~\ref{sec:parameter_staleness} when \textsc{gem} starts to show suboptimal and even divergent behavior (for $\kappa=4$). This indicates that the energy expressed by the proxy is approaching the unknown divergence limit.}
    \label{fig:kappa}
\end{figure}

Depending on the definition of the proxy, the efficiency of the optimization procedure can be increased by adjusting its energy by a constant factor $\kappa$. 
More specifically, to increase the permissible energy in \textsc{gem} proportionally to $\kappa$, we simply redefine the proxy energy to
\begin{equation}
    \label{eq:proxy_def_kappa}
    \mathcal{E}_k(t+1) \triangleq \frac{\kappa}{2}\left(\mu m_{t-\tau_t} + \Delta\theta_t\right)^2.
\end{equation}
From this definition it is clear that $\kappa$ amplifies the energy of the proxy, therefore allowing for larger worker contributions as the range of the proxy energy is extended. As $\pi_t$ depends on the proxy, we can directly include $\kappa$ in $\pi_t$ as 
\begin{equation}
    \label{eq:pi_with_kappa}
    \pi_t = \frac{\sqrt{\kappa}\vert m\vert - \vert\theta - s\vert}{\vert \Delta\theta_t\vert + \epsilon}.
\end{equation}
Since gradually increasing $\kappa$ does not have a significant effect on $m$ (because of $\sqrt{\kappa})$, we modify Equation~\ref{eq:pi_with_kappa} for the purpose of this study to
\begin{equation}
    \pi_t = \frac{\kappa\vert m\vert - \vert\theta - s\vert}{\vert \Delta\theta_t\vert + \epsilon}.
\end{equation}
This effect can be observed from Figure~\ref{fig:kappa}, where the training loss improves as $\kappa$ increases. However, we would like to note that increasing the proxy energy can have adverse effects on the convergence of the training as the energy levels expressed by the proxy is reaching the divergence limit.

\newpage

\begin{sidewaysfigure}[ht]
    \includegraphics[width=\textwidth]{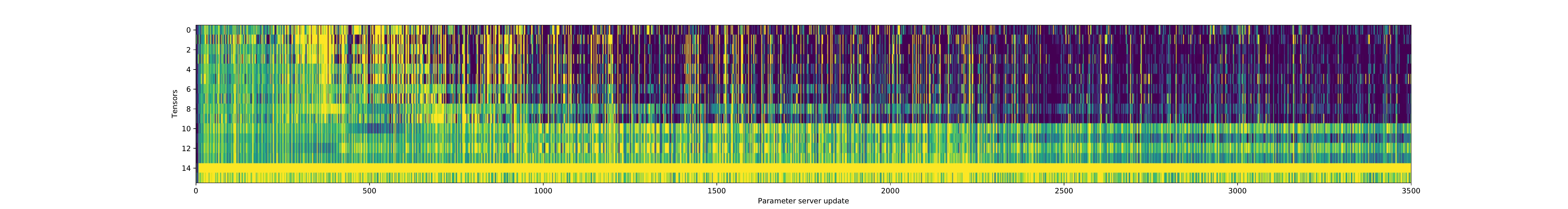}
    \caption{ImageNet experiment, for 8 asynchronous workers. The figure shows how individual AlexNet tensors are adjusted by \textsc{gem}. Individual cells show the median scaling term $\pi_t$ of that particular tensor. Brighter colors indicate large scaling values.}
    \label{fig:alexnet_pi_8}
    \vspace{3cm}
    \includegraphics[width=\textwidth]{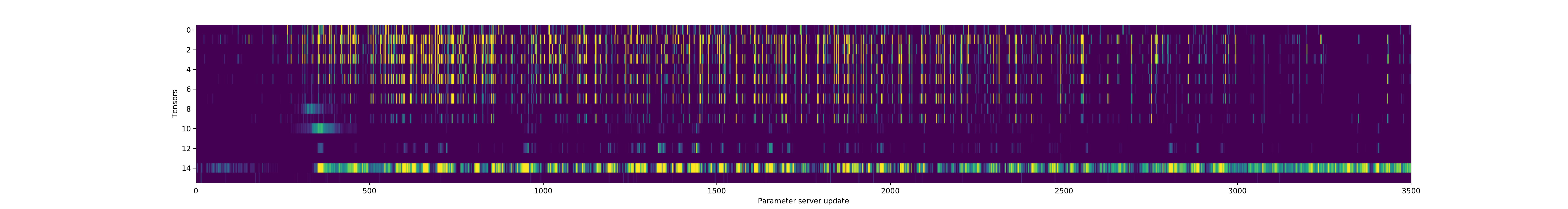}
    \caption{ImageNet experiment, for 16 asynchronous workers. Contrary to Figure~\ref{fig:alexnet_pi_8}, median values for $\pi_t$ are significantly smaller. This is expected as this configuration is obtained by the double amount of workers. Nevertheless, it remains to be investigated.}
    \label{fig:alexnet_pi_16}
\end{sidewaysfigure}

\end{document}